\documentclass{svproc}
\usepackage{amsmath,amssymb,amsfonts}%
\usepackage{graphicx}%
\usepackage{multirow}%
\usepackage{mathrsfs}%
\usepackage[title]{appendix}%
\usepackage{xcolor}%
\usepackage{textcomp}%
\usepackage{manyfoot}%
\usepackage{booktabs}%
\usepackage{algorithm}%
\usepackage{algorithmicx}%
\usepackage{algpseudocode}%
\usepackage{listings}%
\usepackage{cite}
\usepackage{tabularx}
\usepackage{dirtytalk}
\usepackage{xcolor}
\usepackage{hyperref}       
\usepackage{tikz}
\usepackage{subcaption}
\usepackage{comment}
\usepackage{url}

\spnewtheorem{mydefinition}{Definition}{\bfseries}{\normalfont}
\newtheorem{mylemma}{Lemma}
\definecolor{mygreen}{RGB}{34,139,34}
\begin{document}

\mainmatter              
\title{Enhancing GNNs with Architecture-Agnostic Graph Transformations: A Systematic Analysis}
\titlerunning{Enhancing GNNs with Architecture-Agnostic Graph Transformations}  
%

\author{Zhifei Li\inst{1} \and Gerrit Großmann\inst{2} 
\and
Verena Wolf\inst{1,2}}
\authorrunning{Zhifei Li et al.} 

\institute{Saarland Informatics Campus, Saarland University, Saarbrücken, Germany\\
\email{zhli00001@stud.uni-saarland.de},
\and
German Research Center for Artificial Intelligence (DFKI), Kaiserslautern,
Germany
\and
Saarland Informatics Campus,
Saarbrücken,
Germany
}

\maketitle              

\begin{abstract} In recent years, a wide variety of graph neural network (GNN) architectures have emerged, each with its own strengths, weaknesses, and complexities. Various techniques, including rewiring, lifting, and node annotation with centrality values, have been employed as pre-processing steps to enhance GNN performance. However, there are no universally accepted best practices, and the impact of architecture and pre-processing on performance often remains opaque.

This study systematically explores the impact of various graph transformations as pre-processing steps on the performance of common GNN architectures across standard datasets. The models are evaluated based on their ability to distinguish non-isomorphic graphs, referred to as expressivity.

Our findings reveal that certain transformations, particularly those augmenting node features with centrality measures, consistently improve expressivity. However, these gains come with trade-offs, as methods like graph encoding, while enhancing expressivity, introduce numerical inaccuracies widely-used python packages. Additionally, we observe that these pre-processing techniques are limited when addressing complex tasks involving 3-WL and 4-WL indistinguishable graphs.

\end{abstract}
\section{Introduction}\label{sec1}
Applying deep learning techniques to graph-structured data presents unique challenges due to the irregular and heterogeneous nature of graphs compared to the more regular structures found in data like images and text. Molecular graphs, in particular, preserve rich and complex structural information, making them essential for tasks like drug discovery, where understanding molecular interactions and structures is critical for identifying new therapeutic compounds. This complexity makes it difficult to apply traditional deep learning models to graph data, necessitating specialized approaches.

\noindent Graph neural networks (GNNs) have emerged as a powerful solution to these challenges by leveraging the inherent graph structure in their computations \cite{sanchez-lengeling2021a}. Among the various GNN architectures, the Message Passing Neural Network (MPNN) is particularly prominent. MPNNs propagate features from neighboring nodes to a target node via edges, capturing the local neighborhood's structure across multiple layers. The global representation of the entire graph can be obtained by pooling the node representations, often by averaging or summing them \cite{ying2019hierarchical}. Despite extensive research into enhancing the expressivity of MPNNs, particularly in distinguishing non-isomorphic graphs, studies have shown that standard MPNNs are inherently limited by the expressive power of the 1-dimensional Weisfeiler-Leman (1-WL) graph isomorphism heuristic. This limitation has been explored by Morris et al.\ \cite{morris2021weisfeilerlemanneuralhigherorder} and Xu et al.\ \cite{xu2018powerful}, and further analyzed in the context of graph isomorphism distinctiveness by Sato \cite{sato2020survey}. Zhang et al.\ \cite{zhang2023expressive} subsequently categorized efforts to improve GNN expressivity into three main strategies: graph feature enhancement, GNN architecture enhancement, and graph topology enhancement.

\noindent While many approaches focus on modifying graph topology, such as adding or removing edges, to increase expressivity \cite{giraldo2023trade, rong2020dropedge, karhadkar2023fosr, topping2022understanding, gutteridge2023drewdynamicallyrewiredmessage}, these methods often compromise the preservation of isomorphic relationships between graphs. 
Consequently, methods that modify graph topology risk losing the graph's structural integrity, reducing the reliability of downstream applications that depend on consistent graph representations.

\noindent In response to these challenges, our research explores an alternative approach: We systematically study methods that enhance the expressivity of GNN models through graph transformations that preserve isomorphic relationships among graphs. 
The transformations we investigate include the insertion of virtual nodes, centrality-based feature augmentations (such as degree, closeness, betweenness, and eigenvector centrality), distance encoding, graph encoding, subgraph extraction and adding nodes to edges. These methods operate in polynomial time in the number of nodes, offering a promising approach to enhancing GNN expressivity while avoiding the trade-offs commonly associated with topological modifications.

\noindent To evaluate the effectiveness of these graph transformations, we conduct experiments on standard datasets, including the EXP dataset, introduced by Abboud et al.\ \cite{abboud2021surprising} and BREC dataset, introduced by Wang et al.\ \cite{wang2024empiricalstudyrealizedgnn}. Our results demonstrate that certain transformation methods can indeed enhance expressivity, particularly in tasks that are indistinguishable by 1-WL. However, we also observe that these methods face limitations when applied to more complex tasks requiring higher levels of expressivity, such as those evaluated by 3-WL and 4-WL tests. This empirical analysis depends not only on the theoretical properties of the transformations but also on the efficiency and quality of the software libraries used. We employ the DGL library for implementing graph neural networks and NetworkX for graph manipulation, alongside PyTorch as the computational backend. The choice of these libraries influences the scalability, performance, and reliability of our results, which may vary depending on software optimizations and hardware support.

\noindent Our findings demonstrate that graph transformations, particularly node feature augmentation, can significantly enhance GNN expressivity by improving the ability to distinguish non-isomorphic graphs in simpler datasets. However, there is a trade-off: transformations like graph encoding, while boosting expressivity, may also misclassify isomorphic graphs. This highlights the need for further refinement to balance expressivity and accuracy in handling more complex graph structures.

Our manuscript is organised as follows:
In Section \ref{sec2}, we present the foundational background that covers essential definitions and theoretical frameworks that inform our approach. Section \ref{sec3} details the datasets and the graph transformation techniques utilized in our experiments. In Section \ref{sec4}, we outline the experimental setup and provide a comprehensive analysis of the results, showcasing the influence of these transformations on the expressivity of various GNN architectures. Additionally, we discuss the broader implications of our findings, particularly with respect to the challenges posed by higher-order indistinguishability in graph structures. Finally, in Section \ref{sec5}, we summarize our key contributions and offer perspectives on future research directions, emphasizing potential advancements in enhancing GNN expressivity.

\section{Background}\label{sec2}

\subsubsection{Graph Isomorphism} Given two graphs $G=({A}, {X})$ and $G'=({A'}, {X'})$ with adjacency matrices ${A}$ and ${A'}$ and feature matrices ${X}$ and ${X'}$ respectively, an isomorphism between these graphs is a bijective mapping $\pi: \mathcal{V}[G] \longrightarrow \mathcal{V}[G']$ that satisfies ${A}_{uv} = {A'}_{\pi (u) \pi (v)}$ and ${X}_{v} = {X'}_{\pi (v)}$ \cite{zhang2023expressive}. We use $G \cong G'$ to denote that $G$ and $G'$ are isomorphic, where $G'$ can be represented as $\pi (G)$. If there is no bijection $\pi$ that satisfies these conditions, we say that $G$ and $G'$ are non-isomorphic.

\subsubsection{Graph Neural Networks}
A Graph Neural Network (GNN) is a differentiable and learnable function designed to operate on graph-structured data \cite{xu2018powerful}. A GNN that performs a graph classification or regression task is permutation invariant to the ordering or labeling of the nodes, meaning that if the nodes of the input graph are permuted, the output of the GNN remains unchanged. Thus, isomorphic graphs lead to the same GNN output (graph embedding) by design.

\subsubsection{Message Passing Neural Networks (MPNNs)} 
One of the most widely-used paradigms for designing GNNs is the MPNN framework \cite{gilmer2017neural}. This framework consists of multiple layers, where each layer computes a new embedding for a node by

\subsubsection{Input and Computational Graph}
In the context of MPNNs, it is crucial to distinguish between the input graph and the computational graph. The input graph is the foundational structure upon which GNNs perform learning and inference tasks. 
A computational graph \(G^*\) is the graph which is used for message passing derived from the input graph through an explicit or implicit graph transformation.

\subsubsection{Graph Transformation Function}
We consider transformation functions that serve as pre-processing steps, mapping an input graph to a computational graph. Formally, a transformation \( f : G \rightarrow G^* \) must preserve graph isomorphism, meaning that
\( G \cong H \)  implies 
\( f(G) \cong f(H) \).
Additionally, it is crucial that these transformations can be computed efficiently within polynomial time (with respect to the number of nodes).

\subsubsection{Separation Ability}
Separation ability refers to the capacity of a GNN to distinguish between non-isomorphic graphs \cite{morris2021weisfeilerlemanneuralhigherorder,xu2018powerful}. Formally, given two non-isomorphic graphs $G_1 = (A_1, X_1)$ and $G_2 = (A_2, X_2)$, where $X_1 = X_2 = \mathbf{1}$ (i.e., the graphs lack node features), a GNN, $g_\theta(\cdot)$, exhibits separation ability if $g_\theta(G_1) \neq g_\theta(G_2)$, thereby correctly identifying the two graphs as distinct.

This metric is integral to assessing GNN expressivity, as it quantitatively evaluates the model's ability to capture and differentiate structural differences between graphs.

\subsubsection{Weisfeiler-Leman (WL) Test}
Definition and more details about WL Test are provided in Appendix \ref{appendix:WL}.

\section{Methodology}\label{sec3}
\subsection{Datasets}

\subsubsection{EXP Dataset}
The EXP dataset, introduced by Abboud et al.\ \cite{abboud2021surprising}, consists of 600 pairs of non-isomorphic graphs that the 1-WL test fails to distinguish. Each graph is composed of two disconnected components: the \say{core pair,} designed to be non-isomorphic, and the \say{planar component,} which is identical in both graphs to introduce noise.

However, there are only three substantially different core pairs in the dataset, limiting its ability to thoroughly test model expressivity. This lack of diversity has led many recent GNNs to achieve near \(100\%\) accuracy on EXP, making it difficult to use for detailed comparisons. To address this limitation, we also use the BREC dataset, introduced by Wang et al.\ \cite{wang2024empiricalstudyrealizedgnn}.

\subsubsection{BREC Dataset}
The BREC dataset provides a more challenging benchmark with 800 non-isomorphic graphs, organized into 400 pairs across four categories: Basics, Regular, Extension, and CFI. It includes graphs that are indistinguishable by both 1-WL and higher-order tests, such as 3-WL and 4-WL, offering a more rigorous evaluation of model expressivity.

\subsubsection{Task and Dataset Augmentation}
The primary task for both datasets is to distinguish between non-isomorphic graph pairs while correctly identifying isomorphic pairs. To further assess model expressivity, we extended both datasets by adding isomorphic graph pairs. This allows us to evaluate whether the model generates distinct embeddings for non-isomorphic graphs and identical embeddings for isomorphic ones. 
The former measures the expressivity of the GNN, the latter is a sanity check to test the numerical stability of the underlying pipeline.

Samples of these two datasets are shown in Appendix \ref{appendix:datasets}.

\subsection{Graph Transformations}
We evaluate nine graph transformations that preserve graph isomorphism and can be computed within polynomial time, to assess their impact on the performance of standard GNN architectures on the EXP and BREC datasets (\( |V| \) represents the number of nodes and \( |E| \) represents the number of edges.)

\subsubsection{Virtual Node}
Proposed by Gilmer et al.\ \cite{gilmer2017neural}, this method introduces a global virtual node that connects to all other nodes in the graph. This node acts as a shared scratch space for the entire graph, facilitating long-range information exchange during message passing.

In particular, a new node \( v_0 \) is added, and edges \( (v_0, v) \) are created for all \( v \in V(G) \). If two graphs \( G \) and \( H \) are isomorphic, their transformed versions \( G^* \) and \( H^* \) will also be isomorphic, as the isomorphism \( \pi \) extends naturally to the virtual node. The transformation introduces \( O(1) \) additional nodes and \( O(|V|) \) edges, resulting in a time complexity of \( O(|V|) \).

\subsubsection{Centrality-Based Methods}
We consider four centrality measures: degree, closeness, betweenness, and eigenvector centrality. Each centrality measure provides different insights into the structure of a node's neighborhood and its influence in the graph.

\begin{itemize}
    \item [\textbullet] \textbf{Degree Centrality:} Computes the number of edges incident to each node. This method can be computed in \( O(|V| + |E|) \).
    \item [\textbullet] \textbf{Closeness Centrality:} Measures how close a node is to all others in the graph, based on shortest path distances. For unweighted graphs, the complexity is \( O(|V|^2) \) for sparse graphs.
    \item [\textbullet] \textbf{Betweenness Centrality:} Quantifies the number of shortest paths passing through a node, with a time complexity of \( O(|V| \cdot |E|) \), making it more computationally expensive for larger graphs.
    \item [\textbullet] \textbf{Eigenvector Centrality:} Evaluates the influence of a node based on the importance of its neighbors. Efficient approximation methods yield a time complexity of \( O(|E|) \).
\end{itemize}

\subsubsection{Distance Encoding}

Distance encoding augments node features by incorporating information about shortest path distances between nodes. This method enhances the graph's structural representation by introducing relative distance information as node features. The shortest path distances are computed using BFS, leading to a complexity of \( O(|V|(|V| + |E|))\).

\subsubsection{Graph Encoding with Laplacian Eigenvectors}
Inspired by Dwivedi et al.\ \cite{dwivedi2022benchmarkinggraphneuralnetworks}, this method uses the top-k eigenvectors of the graph Laplacian to capture global structural information. These eigenvectors highlight topological features such as clusters and connectivity patterns. The computation of the eigenvectors through eigenvalue decomposition has a time complexity of \( O(|V|^3) \), making it suitable for smaller graphs.

\subsubsection{Subgraph Extraction}

Following Zhao et al.\ \cite{zhao2022starssubgraphsupliftinggnn}, we extract subgraphs (specifically, \say{ego graphs}) for each node by considering nodes within a fixed radius (e.g., 2 hops). The size of these subgraphs is then used as additional node features.
It takes \(O(|V| + |E|)\) to extract an ego graph for a single node. For each node in the graph, the total complexity is \(O(|V|(|V| + |E|))\). 

\subsubsection{Add Extra Node on Each Edge}

This transformation adds an intermediary node on each edge, turning direct connections into two-step paths. This modification allows GNNs to capture more fine-grained interactions and reduces the over-squashing problem. Specifically, for each edge \( (u, v) \), we introduce a new node \( w \) and create edges \( (u, w) \) and \( (w, v) \). The transformation preserves isomorphisms since any isomorphism \( \pi \) between \( G \) and \( H \) extends to their modified versions \( G^* \) and \( H^* \). The complexity of this method is \( O(|E|) \).

\begin{figure}[h!]
    \centering
    \begin{subfigure}[b]{0.3\textwidth}
        \centering
        \includegraphics[width=\textwidth]{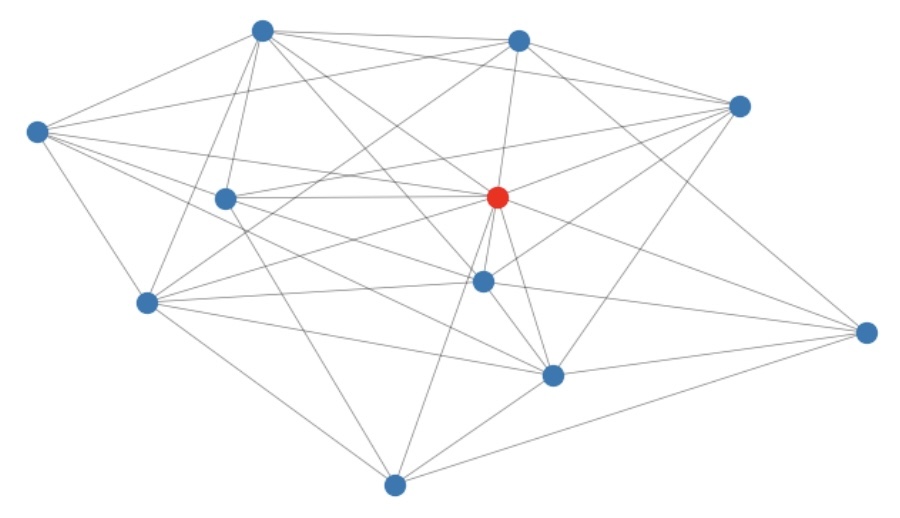}
        \caption{Virtual Node}
        \label{fig:sub1}
    \end{subfigure}
    \hfill
    \begin{subfigure}[b]{0.3\textwidth}
        \centering
        \includegraphics[width=\textwidth]{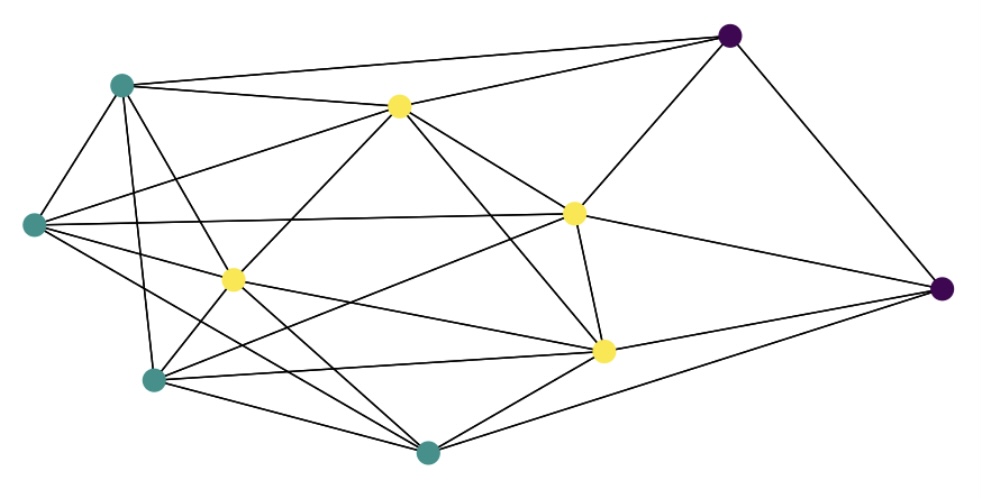}
        \caption{Degree Centrality}
        \label{fig:sub1x}
    \end{subfigure}
    \hfill
    \begin{subfigure}[b]{0.3\textwidth}
        \centering
        \includegraphics[width=\textwidth]{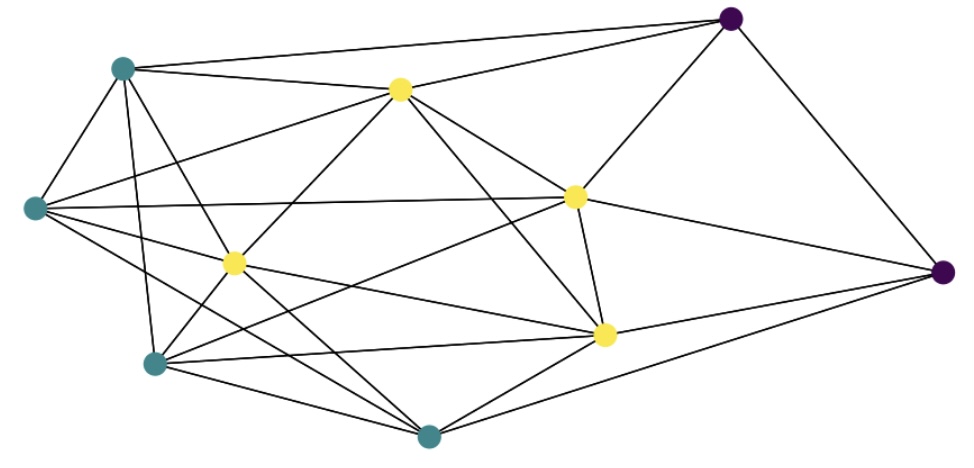}
        \caption{Closeness Centrality}
        \label{fig:sub2}
    \end{subfigure}

    \begin{subfigure}[b]{0.32\textwidth}
        \centering
        \includegraphics[width=\textwidth]{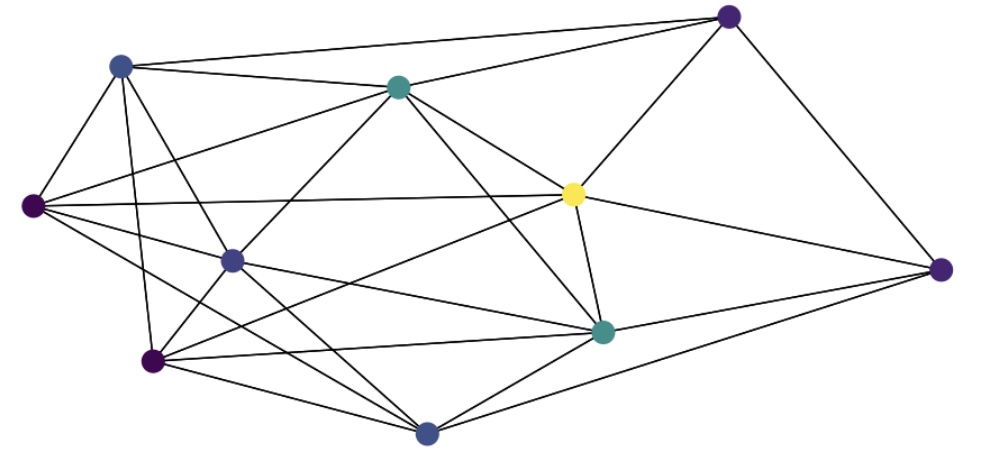}
        \caption{Betweenness Centrality}
        \label{fig:sub3}
    \end{subfigure}
    \hfill
    \begin{subfigure}[b]{0.3\textwidth}
        \centering
        \includegraphics[width=\textwidth]{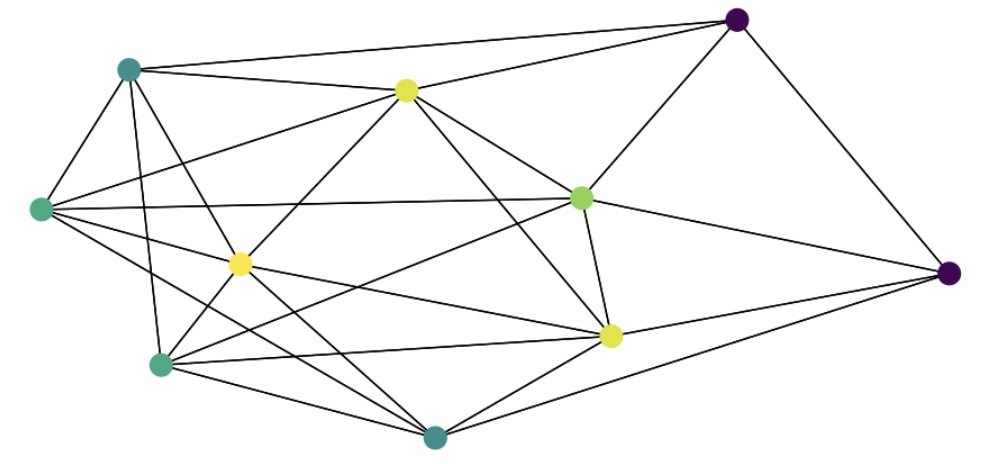}
        \caption{Eigenvector Centrality}
        \label{fig:sub4}
    \end{subfigure}
    \hfill
    \begin{subfigure}[b]{0.3\textwidth}
        \centering
        \includegraphics[width=\textwidth]{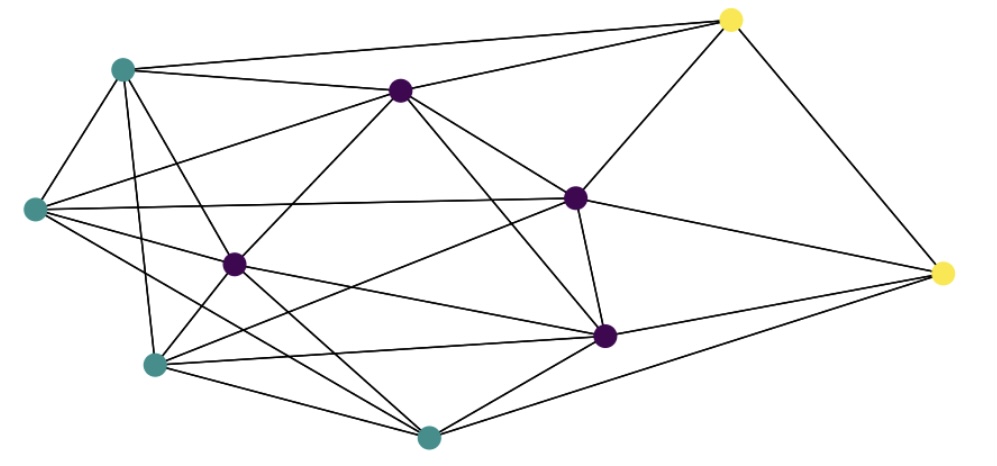}
        \caption{Distance Encoding}
        \label{fig:sub5}
    \end{subfigure}

    \begin{subfigure}[b]{0.3\textwidth}
        \centering
        \includegraphics[width=\textwidth]{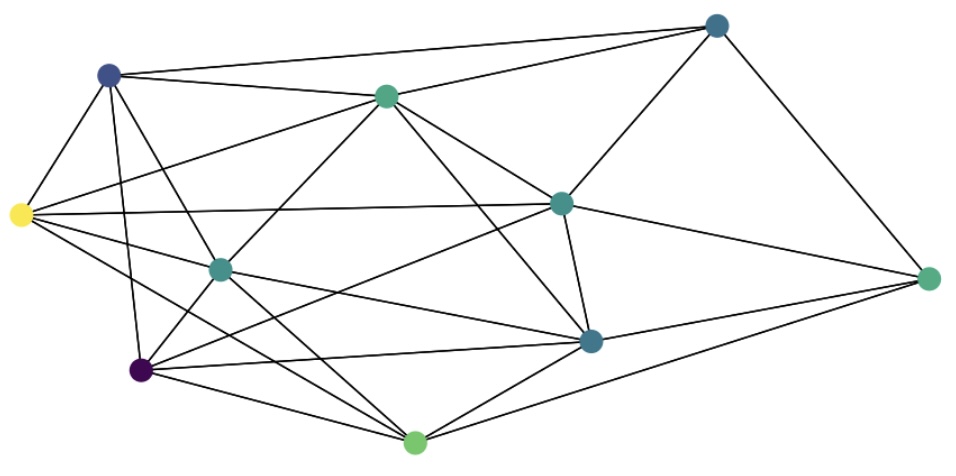}
        \caption{Graph Encoding}
        \label{fig:sub6}
    \end{subfigure}
    \hfill
    \begin{subfigure}[b]{0.3\textwidth}
        \centering
        \includegraphics[width=\textwidth]{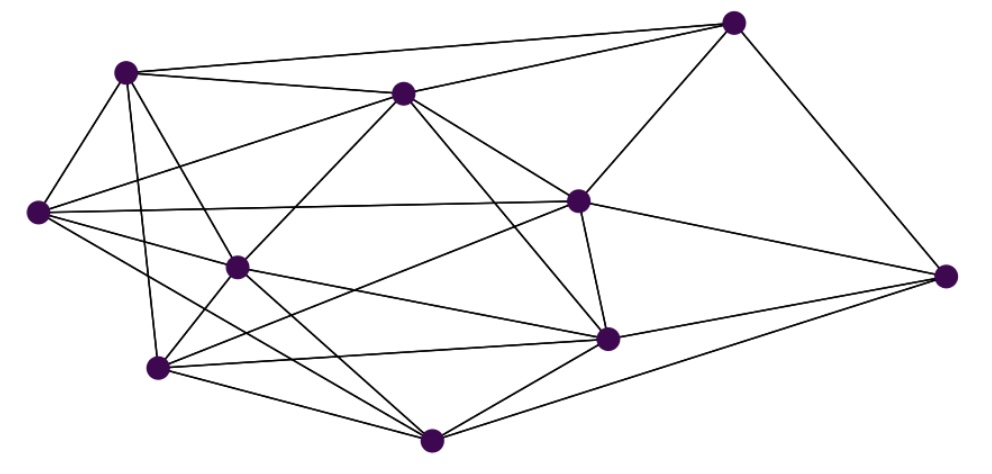}
        \caption{Subgraph Extraction}
        \label{fig:sub7}
    \end{subfigure}
    \hfill
    \begin{subfigure}[b]{0.3\textwidth}
        \centering
        \includegraphics[width=\textwidth]{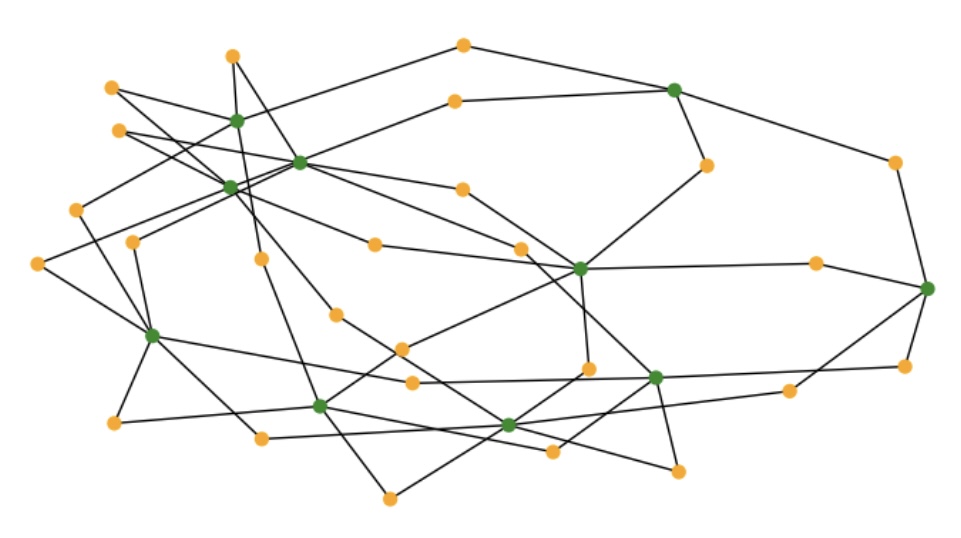}
        \caption{Extra Node}
        \label{fig:sub8}
    \end{subfigure}

\caption{Transformation Example. For Virtual Node method, \textcolor{red}{red} indicates an added virtual node. For Extra Node method, \textcolor{orange}{orange} indicates an added extra node, and \textcolor{mygreen}{green} represents the original node.}

    \label{fig:trans}
\end{figure}

The examples of applying these transformation methods are shown in Figure \ref{fig:trans}.

\subsection{Evaluation Methodology}
To assess the expressivity of GNN models, we utilize three key metrics: \textbf{Equivalence Class Count (ECC)}, \textbf{False Positives (FP)}, and \textbf{False Negatives (FN)}. These metrics offer a comprehensive evaluation of how effectively the models differentiate between isomorphic and non-isomorphic graphs.

\subsubsection{Equivalence Class Count (ECC)} The ECC metric measures how well the model distinguishes between graphs by counting the number of unique graph embeddings produced by the GNN. Specifically, if two graphs \(G_1\) and \(G_2\) produce different embeddings, they are assigned to different equivalence classes. The total number of such classes reflects the model’s ability to distinguish non-isomorphic graphs while ideally grouping isomorphic graphs into the same class. This metric not only captures the model’s expressivity but also highlights potential errors, such as incorrectly separating isomorphic graphs or failing to differentiate non-isomorphic ones.

\subsubsection{False Positive (FP):} This represents the number of isomorphic graph pairs that the model incorrectly classifies as non-isomorphic. In other words, the model mistakenly identifies two isomorphic graphs as different.

\subsubsection{False Negative (FN):} This measures the number of non-isomorphic graph pairs that the model incorrectly classifies as isomorphic, meaning the model mistakenly identifies two non-isomorphic graphs as identical.

\section{Experiment}\label{sec4}
\subsubsection{Experimental Setup}

We evaluate the effect of the graph transformation methods described in this paper by applying them as pre-processing steps to graphs from EXP and BREC datasets. For each dataset, we employ three distinct GNN architectures to assess the impact of these transformations on model performance: Graph Isomorphism Networks (GIN) \cite{xu2018powerful}, Principal Neighbourhood Aggregation (PNA) \cite{corso2020principalneighbourhoodaggregationgraph}, and Deep Set (DS) \cite{zaheer2018deepsets, grossmann2023poster}. 
GIN relies on message passing and are as powerful as the 1-WL test for graph isomorphism. PNA extends GIN by using multiple aggregators and degree scalers to capture a broader range of structural information. In contrast, Deep Set operates on node features in a permutation-invariant manner, making it distinct from typical message-passing models by directly learning graph representations through aggregation. A more detailed description of these architectures is provided in Appendix \ref{appendix:GNNs} and further details regarding the experimental procedures can be found in the Appendix \ref{appendix:setup}.

\subsubsection{Results}
The results for the EXP and BREC datasets are summarized in Table \ref{tab:EXP} and \ref{tab:BREC}, respectively. Table \ref{tab:BREC-sub} shows more detailed results of BREC dataset.

\vspace{-1em}
\begin{table}[h!]
\centering
\begin{scriptsize}
\renewcommand{\arraystretch}{1.2}
\scalebox{0.95}{
\begin{tabular}
{l|ccc|ccc|ccc|ccc}
\toprule
\multirow{3}{*}{\textbf{Method}} & \multicolumn{6}{c|}{\textbf{EXP-original}} & \multicolumn{6}{c}{\textbf{EXP-modified}} \\ 
\cline{2-13}
& \multicolumn{3}{c|}{\textbf{ECC}} & \multicolumn{3}{c|}{\textbf{FN}} & \multicolumn{3}{c|}{\textbf{ECC}} & \multicolumn{3}{c}{\textbf{FP}} \\
\cline{2-13}
 & \textbf{GIN} & \textbf{PNA} & \textbf{DS} & \textbf{GIN} & \textbf{PNA} & \textbf{DS} & \textbf{GIN} & \textbf{PNA} & \textbf{DS} & \textbf{GIN} & \textbf{PNA} & \textbf{DS} \\ 
\midrule
Base
&599  &593  &54  &600  &600  &600 &599  &546  &50 &0  &0  &0 \\
\midrule
Virtual Node
&623  &500  &56  &600  &600  &600 &685  &491  &48 &0  &0  &0 \\
\midrule
Degree
&598  &601  &67  &600  &600  &600  &474  &573  &82 &0  &0  &0\\
\midrule
Closeness
&1193  &519  &84 &151  &279  &600  &1190  &1010  &131 &0  &0  &0 \\
\midrule
Betweenness
&1194  &\textcolor{orange}{1145}  &44 & 175  &255  &600  &1182  &1139  &75 &0  &0  &0 \\
\midrule
Eigenvector 
&599  &587  &\textcolor{orange}{127} &600  &600  &600 &598  &515  &48 &0  &0  &0 \\
\midrule
Distance Encoding
&1188  &1108  &\textcolor{mygreen}{232} &\textcolor{orange}{33}  & \textcolor{orange}{200}  &600  & \textcolor{mygreen}{1197}  &940  &183 &0  &0  &0 \\
\midrule
Graph Encoding
& \textcolor{orange}{1195}  & \textcolor{mygreen}{1174}  &71 &324  &382  &600  &\textcolor{red!70!black}{1745}  &\textcolor{red!70!black}{1555}  &93  &\textcolor{red!70!black}{26} &0 &0\\
\midrule
Subgraph Extraction
& \textcolor{mygreen}{1198}  &1110  &123 & \textcolor{mygreen}{0}  & \textcolor{mygreen}{42}  &600 &\textcolor{orange}{1195}  &1037  &171 &0  &0  &0\\
\midrule
Extra Node
&601  &589  &9 &600  &600  &600 &600  &602  &75 &0  &0  &0\\
\bottomrule
\end{tabular}
}
\end{scriptsize}
\caption{Comparison of methods on original EXP dataset and modified EXP dataset with 600 added isomorphic pairs. For ECC under original dataset, higher values indicate better performance, whereas under modified dataset, we expect small changes. Lower values are preferred for FN and FP.
\textcolor{mygreen}{green} - best results; \textcolor{orange}{orange} - second best results \textcolor{red!70!black}{red} - noteworthy results. 
}
\label{tab:EXP}
\end{table}

\begin{table}[hbtp]
\centering
\begin{scriptsize}
\renewcommand{\arraystretch}{1.2}
\scalebox{0.95}{
\begin{tabular}
{l|ccc|ccc|ccc|ccc}
\toprule
\multirow{3}{*}{\textbf{Method}} & \multicolumn{6}{c|}{\textbf{BREC-original}} & \multicolumn{6}{c}{\textbf{BREC-modified}} \\ 
\cline{2-13}
& \multicolumn{3}{c|}{\textbf{ECC}} & \multicolumn{3}{c|}{\textbf{FN}} & \multicolumn{3}{c|}{\textbf{ECC}} & \multicolumn{3}{c}{\textbf{FP}} \\
\cline{2-13}
 & \textbf{GIN} & \textbf{PNA} & \textbf{DS} & \textbf{GIN} & \textbf{PNA} & \textbf{DS} & \textbf{GIN} & \textbf{PNA} & \textbf{DS} & \textbf{GIN} & \textbf{PNA} & \textbf{DS} \\ 
\midrule
Base
&294  &401  &70  &400  &400  &400  &297  &418  &70 &0  &0  &0 \\
\midrule
Virtual Node
&286  &258  &70  &400  &400  &400  &293  &263  &70 &0  &0  &0 \\
\midrule
Degree
&308  &399  &125  &400  &400  &400  &297  &415  &113 &0  &0  &0\\
\midrule
Closeness
&382  &488  &197 &347  &\textcolor{orange}{333}  &400  &384  &515  &192 &0  &1  &0 \\
\midrule
Betweenness
& \textcolor{orange}{502}  &\textcolor{mygreen}{600}  &148 &342  &\textcolor{mygreen}{310}  &400  &514  &605  &151 &0  &0  &0 \\
\midrule
Eigenvector 
&312  &408  &105 &397  &400  &400 &295  &429  &110 &0  &0  &0 \\
\midrule
Distance Encoding
&395  &487  &203 &349  &343  &400  &379  &509  &185 &0  &0  &0 \\
\midrule
\textcolor{red!70!black}{Graph Encoding}
&\textcolor{red!70!black}{770}  &\textcolor{red!70!black}{772}  &326 &\textcolor{mygreen}{292}  &361  &\textcolor{mygreen}{399}  &\textcolor{red!70!black}{1149}  &\textcolor{red!70!black}{1142}  &371  &\textcolor{red!70!black}{30} &0 &1\\
\midrule
Subgraph Extraction
&373  &557  &245 &359  &340  &400 &379  &543  &250 &0  &0  &0\\
\midrule
Extra Node
& \textcolor{mygreen}{578}  &\textcolor{orange}{594}  &110 &\textcolor{orange}{298}  &369  &400  &580  &586  &128 &0  &0  &0\\
\bottomrule
\end{tabular}
}
\end{scriptsize}
\caption{Comparison of methods on the original and modified BREC datasets (with 400 added isomorphic pairs). For ECC under original dataset, higher values indicate better performance, whereas under modified dataset, we expect small changes. Lower values are preferred for FN and FP.
\textcolor{mygreen}{green}: best results; \textcolor{orange}{orange}: second-best results; \textcolor{red!70!black}{red}: noteworthy results.}
\label{tab:BREC}
\end{table}

\begin{table}[h!]
\centering
\begin{scriptsize}
\renewcommand{\arraystretch}{1.2} 
\scalebox{0.8}{
\begin{tabular}
{l|cccccc|cccccc|cccccc|cccccc}
\toprule
\multirow{3}{*}{\textbf{Method}} & \multicolumn{6}{c|}{\textbf{Basics(60)}} & \multicolumn{6}{c|}{\textbf{Regular(140)}} & \multicolumn{6}{c|}{\textbf{Extension(100)}} & \multicolumn{6}{c}{\textbf{CFI(100)}} \\ 
\cline{2-25}
& \multicolumn{2}{c}{\textbf{GIN}} & \multicolumn{2}{c}{\textbf{PNA}} & \multicolumn{2}{c|}{\textbf{DS}} & \multicolumn{2}{c}{\textbf{GIN}} & \multicolumn{2}{c}{\textbf{PNA}} & \multicolumn{2}{c}{\textbf{DS}}& \multicolumn{2}{c}{\textbf{GIN}} & \multicolumn{2}{c}{\textbf{PNA}} & \multicolumn{2}{c}{\textbf{DS}}& \multicolumn{2}{c}{\textbf{GIN}} & \multicolumn{2}{c}{\textbf{PNA}} & \multicolumn{2}{c}{\textbf{DS}} \\
\cline{2-25}
 & \textbf{FN} & \textbf{FP} & \textbf{FN} & \textbf{FP} & \textbf{FN} & \textbf{FP} & \textbf{FN} & \textbf{FP} & \textbf{FN} & \textbf{FP} & \textbf{FN} & \textbf{FP} & \textbf{FN} & \textbf{FP} & \textbf{FN} & \textbf{FP} & \textbf{FN} & \textbf{FP} &
 \textbf{FN} & \textbf{FP} & \textbf{FN} & \textbf{FP} & \textbf{FN} & \textbf{FP} \\ 
\midrule
Base
&60  &0  &60  &0  &60  &0  &140  &0  &140  &0  &140  &0
&100  &0  &100  &0   &100  &0   &100  &0   &100  &0   &100  &0\\
\midrule
Virtual Node
&60  &0  &60  &0  &60  &0 &140  &0  &140  &0  &140  &0 
&100  &0   &100  &0   &100 &0  &100  &0   &100  &0   &100  &0\\
\midrule
Degree
&60  &0  &60  &0  &60  &0 &140  &0  &140  &0  &140  &0
&100  &0   &100  &0   &100  &0   &100  &0   &100  &0   &100  &0\\
\midrule
Closeness
&45  &0  &45  &0  &60  &0 &140  &0  &\textcolor{orange}{128} &1  &140  &0
&\textcolor{orange}{63}  &0   &\textcolor{orange}{60}  &0  &100  &0   &\textcolor{orange}{99}  &0   &100  &0   &100  &0\\
\midrule
Betweenness
&42  &0  &\textcolor{mygreen}{40}  &0  &60  &0 &140  &0  &\textcolor{mygreen}{116} &0  &140  &0 
&\textcolor{orange}{63}  &0   &\textcolor{mygreen}{56}  &0  &100  &0   &\textcolor{mygreen}{97}  &0   &\textcolor{orange}{98}  &0   &100  &0\\
\midrule
Eigenvector 
&60  &0  &60  &0  &60  &0 &140  &0  &140 &0  &140  &0 
&100  &0   &100  &0   &100  &0   &\textcolor{mygreen}{97}  &0   &100  &0   &100  &0\\
\midrule
Distance Encoding
&46  &0  &\textcolor{orange}{44}  &0  &60  &0 &137 &0 &133 &0  &140  &0
&69  &0   &69  &0  &100  &0   &\textcolor{mygreen}{97}  &0   &\textcolor{mygreen}{97}  &0   &100  &0\\
\midrule
Graph Encoding
&\textcolor{mygreen}{19}  &\textcolor{red!70!black}{12}  &\textcolor{mygreen}{40}  &0  &\textcolor{mygreen}{59}  &\textcolor{red!70!black}{1} &\textcolor{orange}{124}  &\textcolor{red!70!black}{12}  &139 &0  &140  &0
&\textcolor{mygreen}{49}  &6   &82  &0  &100  &0   &100  &0   &100  &0   &100  &0\\
\midrule
Subgraph Extraction
&46  &0  &\textcolor{orange}{44}  &0  &60  &0 &140  &0  &135 &0  &140  &0
&73  &0   &61  &0  &100  &0   &100  &0   &100  &0   &100  &0\\
\midrule
Extra Node
&\textcolor{orange}{38}  &0  &48  &0  &60  &0 &\textcolor{mygreen}{86}  &0  &139 &0  &140  &0
&74  &0   &82  &0  &100  &0   &100  &0   &100  &0   &100  &0\\
\bottomrule
\end{tabular}
}
\end{scriptsize}
\caption{Comparison of methods on each sub-dataset of BREC dataset. Lower is better.
\textcolor{mygreen}{green}: best results; \textcolor{orange}{orange}: second best results;
\textcolor{red!70!black}{red} - important results that needs to be noticed.
}
\label{tab:BREC-sub}
\end{table}

\vspace{-3em}

\subsubsection{Observation 1:} \textbf{Enhanced Expressivity Through Pre-Processing}

\noindent 
Table \ref{tab:EXP} and \ref{tab:BREC} reveal that the expressivity of GNN models, especially GIN and PNA, can be significantly enhanced by applying various graph transformation techniques as pre-processing steps. In particular, methods involving node feature augmentation consistently improve the models' ability to differentiate between non-isomorphic graphs. This insight underscores the potential of node feature augmentation to serve as a powerful tool in refining GNN architectures.

\subsubsection{Observation 2:} \textbf{Challenges with Higher-Order Indistinguishability}

Despite the clear gains in expressivity achieved through pre-processing, the results also reveal a critical limitation: these enhancements fall short when addressing more complex problems, such as those involving 3-WL and 4-WL indistinguishable graphs, as shown under CFI sub-dataset in Table \ref{tab:BREC-sub}. This observation indicates that while pre-processing techniques can extend the expressivity of GNNs, they are not applicable to all graph-related challenges. The persistence of these limitations suggests that more advanced (i.e., computationally costly) methods may be required to overcome the expressivity barriers posed by higher-order graph structures. 

\subsubsection{Observation 3:} \textbf{Graph Encoding’s Impact on Expressivity-Equivalence Trade-off}
A critical observation from Table \ref{tab:EXP}, \ref{tab:BREC}, \ref{tab:BREC-sub} involves the trade-off between distinguishing non-isomorphic pairs and preserving structural equivalence for isomorphic pairs. Graph encoding methods show an increased ability to differentiate non-isomorphic graphs, improving the Equivalence Class Count (ECC). However, this enhanced expressivity comes at the cost of misclassifying isomorphic pairs, resulting in a higher number of false positives than other transformation methods (noteworthy results shown in \textcolor{red!70!black}{red}).

We initially incorporated isomorphic graphs into the training data as a sanity check. However, we subsequently observed that discrepancies in the forward pass of the GNN resulted in different embeddings for isomorphic graphs. This behavior was consistently observed across both PyG and DGL, even when the input values were properly rounded to a specified number of decimal places.


\section{Conclusion and Future Work}\label{sec5}
This study examined the role of graph transformations as preprocessing steps to enhance the expressivity of GNNs, with a focus on preserving graph isomorphism. We showed that these transformations, particularly node feature augmentation, can significantly boost the expressivity of simple GNN architectures while keeping computations within polynomial time.

However, our findings also highlight some limitations. While these transformations improve expressivity, they struggle with more complex tasks, especially when dealing with graphs that remain indistinguishable by the 3-WL and 4-WL tests. This suggests that preprocessing alone may not be sufficient and may need to be combined with more sophisticated GNN architectures to address higher-order graph structures. Another key challenge is the trade-off observed with transformations like graph encoding: while they enhance the model's ability to distinguish non-isomorphic graphs, they simultaneously increase the likelihood of misclassifying isomorphic pairs. This trade-off underscores the need to balance expressivity with robustness, pointing to future research that refines these methods or explores hybrid approaches to address both challenges.

Moving forward, future work could involve combining these preprocessing techniques with advanced GNN models or assessing the benefits of integrating multiple transformations. Comparative studies between graph transformation methods and topology modification approaches could also offer valuable insights. Lastly, applying these methods to large-scale, real-world datasets will be critical for validating their practical effectiveness.

\section*{Acknowledgments}
We thank Rebekka Burkholz for her valuable comments on the manuscript.

%
%
\bibliographystyle{unsrt}
\bibliography{cite}

\begin{thebibliography}{10}

\bibitem{sanchez-lengeling2021a}
Benjamin Sanchez-Lengeling, Emily Reif, Adam Pearce, and Alexander~B. Wiltschko.
\newblock A gentle introduction to graph neural networks.
\newblock {\em Distill}, 2021.
\newblock https://distill.pub/2021/gnn-intro.

\bibitem{ying2019hierarchical}
Rex Ying, Jiaxuan You, Christopher Morris, Xiang Ren, William~L. Hamilton, and Jure Leskovec.
\newblock Hierarchical graph representation learning with differentiable pooling, 2019.

\bibitem{morris2021weisfeilerlemanneuralhigherorder}
Christopher Morris, Martin Ritzert, Matthias Fey, William~L. Hamilton, Jan~Eric Lenssen, Gaurav Rattan, and Martin Grohe.
\newblock Weisfeiler and leman go neural: Higher-order graph neural networks, 2021.

\bibitem{xu2018powerful}
Keyulu Xu, Weihua Hu, Jure Leskovec, and Stefanie Jegelka.
\newblock How powerful are graph neural networks?
\newblock {\em arXiv preprint arXiv:1810.00826}, 2018.

\bibitem{sato2020survey}
Ryoma Sato.
\newblock A survey on the expressive power of graph neural networks.
\newblock {\em arXiv preprint arXiv:2003.04078}, 2020.

\bibitem{zhang2023expressive}
Bingxu Zhang, Changjun Fan, Shixuan Liu, Kuihua Huang, Xiang Zhao, Jincai Huang, and Zhong Liu.
\newblock The expressive power of graph neural networks: A survey, 2023.

\bibitem{giraldo2023trade}
Jhony~H Giraldo, Konstantinos Skianis, Thierry Bouwmans, and Fragkiskos~D Malliaros.
\newblock On the trade-off between over-smoothing and over-squashing in deep graph neural networks.
\newblock In {\em ACM International Conference on Information and Knowledge Management}, 2023.

\bibitem{rong2020dropedge}
Yu~Rong, Wenbing Huang, Tingyang Xu, and Junzhou Huang.
\newblock Dropedge: Towards deep graph convolutional networks on node classification, 2020.

\bibitem{karhadkar2023fosr}
Kedar Karhadkar, Pradeep~Kr. Banerjee, and Guido Montúfar.
\newblock Fosr: First-order spectral rewiring for addressing oversquashing in gnns, 2023.

\bibitem{topping2022understanding}
Jake Topping, Francesco~Di Giovanni, Benjamin~Paul Chamberlain, Xiaowen Dong, and Michael~M. Bronstein.
\newblock Understanding over-squashing and bottlenecks on graphs via curvature, 2022.

\bibitem{gutteridge2023drewdynamicallyrewiredmessage}
Benjamin Gutteridge, Xiaowen Dong, Michael Bronstein, and Francesco~Di Giovanni.
\newblock Drew: Dynamically rewired message passing with delay, 2023.

\bibitem{abboud2021surprising}
Ralph Abboud, İsmail~İlkan Ceylan, Martin Grohe, and Thomas Lukasiewicz.
\newblock The surprising power of graph neural networks with random node initialization, 2021.

\bibitem{wang2024empiricalstudyrealizedgnn}
Yanbo Wang and Muhan Zhang.
\newblock An empirical study of realized gnn expressiveness, 2024.

\bibitem{gilmer2017neural}
Justin Gilmer, Samuel~S. Schoenholz, Patrick~F. Riley, Oriol Vinyals, and George~E. Dahl.
\newblock Neural message passing for quantum chemistry, 2017.

\bibitem{dwivedi2022benchmarkinggraphneuralnetworks}
Vijay~Prakash Dwivedi, Chaitanya~K. Joshi, Anh~Tuan Luu, Thomas Laurent, Yoshua Bengio, and Xavier Bresson.
\newblock Benchmarking graph neural networks, 2022.

\bibitem{zhao2022starssubgraphsupliftinggnn}
Lingxiao Zhao, Wei Jin, Leman Akoglu, and Neil Shah.
\newblock From stars to subgraphs: Uplifting any gnn with local structure awareness, 2022.

\bibitem{corso2020principalneighbourhoodaggregationgraph}
Gabriele Corso, Luca Cavalleri, Dominique Beaini, Pietro Liò, and Petar Veličković.
\newblock Principal neighbourhood aggregation for graph nets, 2020.

\bibitem{zaheer2018deepsets}
Manzil Zaheer, Satwik Kottur, Siamak Ravanbakhsh, Barnabas Poczos, Ruslan Salakhutdinov, and Alexander Smola.
\newblock Deep sets, 2018.

\bibitem{grossmann2023poster}
G.~Großmann.
\newblock Deep sets are viable graph learners, complex networks conference, 2023.

\bibitem{vaswani2023attentionneed}
Ashish Vaswani, Noam Shazeer, Niki Parmar, Jakob Uszkoreit, Llion Jones, Aidan~N. Gomez, Lukasz Kaiser, and Illia Polosukhin.
\newblock Attention is all you need, 2023.

\bibitem{kim2022puretransformerspowerfulgraph}
Jinwoo Kim, Tien~Dat Nguyen, Seonwoo Min, Sungjun Cho, Moontae Lee, Honglak Lee, and Seunghoon Hong.
\newblock Pure transformers are powerful graph learners, 2022.

\bibitem{feng2022powerful}
Jiarui Feng, Yixin Chen, Fuhai Li, Anindya Sarkar, and Muhan Zhang.
\newblock How powerful are k-hop message passing graph neural networks.
\newblock {\em Advances in Neural Information Processing Systems}, 35:4776--4790, 2022.

\bibitem{veličković2022message}
Petar Veličković.
\newblock Message passing all the way up, 2022.

\bibitem{balcilar2021breaking}
Muhammet Balcilar, Pierre Héroux, Benoit Gaüzère, Pascal Vasseur, Sébastien Adam, and Paul Honeine.
\newblock Breaking the limits of message passing graph neural networks, 2021.

\bibitem{GNNBook-ch12-guo}
Xiaojie Guo, Shiyu Wang, and Liang Zhao.
\newblock Graph neural networks: Graph transformation.
\newblock In Lingfei Wu, Peng Cui, Jian Pei, and Liang Zhao, editors, {\em Graph Neural Networks: Foundations, Frontiers, and Applications}, pages 251--275. Springer Singapore, Singapore, 2022.

\bibitem{weisfeiler1968reduction}
Boris Weisfeiler and Andrei Leman.
\newblock The reduction of a graph to canonical form and the algebra which appears therein.
\newblock {\em nti, Series}, 2(9):12--16, 1968.

\bibitem{Huang_2021}
Ningyuan~Teresa Huang and Soledad Villar.
\newblock A short tutorial on the weisfeiler-lehman test and its variants.
\newblock In {\em ICASSP 2021 - 2021 IEEE International Conference on Acoustics, Speech and Signal Processing (ICASSP)}. IEEE, June 2021.

\bibitem{nikolentzos2020k}
Giannis Nikolentzos, George Dasoulas, and Michalis Vazirgiannis.
\newblock k-hop graph neural networks.
\newblock {\em Neural Networks}, 130:195--205, 2020.

\bibitem{maron2020provably}
Haggai Maron, Heli Ben-Hamu, Hadar Serviansky, and Yaron Lipman.
\newblock Provably powerful graph networks, 2020.

\bibitem{NEURIPS2023_ebf95a6f}
Fabian Jogl, Maximilian Thiessen, and Thomas G\"{a}rtner.
\newblock Expressivity-preserving gnn simulation.
\newblock In A.~Oh, T.~Naumann, A.~Globerson, K.~Saenko, M.~Hardt, and S.~Levine, editors, {\em Advances in Neural Information Processing Systems}, volume~36, pages 74544--74581. Curran Associates, Inc., 2023.

\bibitem{hwang2022an}
EunJeong Hwang, Veronika Thost, Shib~Sankar Dasgupta, and Tengfei Ma.
\newblock An analysis of virtual nodes in graph neural networks for link prediction (extended abstract).
\newblock In {\em The First Learning on Graphs Conference}, 2022.

\bibitem{zhang2024rethinking}
Bohang Zhang, Shengjie Luo, Liwei Wang, and Di~He.
\newblock Rethinking the expressive power of gnns via graph biconnectivity, 2024.

\bibitem{GNNBook-ch5-li}
Pan Li and Jure Leskovec.
\newblock The expressive power of graph neural networks.
\newblock In Lingfei Wu, Peng Cui, Jian Pei, and Liang Zhao, editors, {\em Graph Neural Networks: Foundations, Frontiers, and Applications}, pages 63--98. Springer Singapore, Singapore, 2022.

\bibitem{tang2023chebnet}
Shanshan Tang, Bo~Li, and Haijun Yu.
\newblock Chebnet: Efficient and stable constructions of deep neural networks with rectified power units via chebyshev approximations, 2023.

\bibitem{thakoor2023largescalerepresentationlearninggraphs}
Shantanu Thakoor, Corentin Tallec, Mohammad~Gheshlaghi Azar, Mehdi Azabou, Eva~L. Dyer, Rémi Munos, Petar Veličković, and Michal Valko.
\newblock Large-scale representation learning on graphs via bootstrapping, 2023.

\bibitem{papp2022theoreticalcomparisongraphneural}
Pál~András Papp and Roger Wattenhofer.
\newblock A theoretical comparison of graph neural network extensions, 2022.

\bibitem{li2023distancematrixgeometricdeep}
Zian Li, Xiyuan Wang, Yinan Huang, and Muhan Zhang.
\newblock Is distance matrix enough for geometric deep learning?, 2023.

\end{thebibliography}

\nocite{vaswani2023attentionneed}
\nocite{kim2022puretransformerspowerfulgraph}
\nocite{sanchez-lengeling2021a}
\nocite{ying2019hierarchical}
\nocite{morris2021weisfeilerlemanneuralhigherorder}
\nocite{xu2018powerful}
\nocite{sato2020survey}
\nocite{zhang2023expressive}
\nocite{feng2022powerful}
\nocite{giraldo2023trade}
\nocite{karhadkar2023fosr}
\nocite{topping2022understanding}
\nocite{gutteridge2023drewdynamicallyrewiredmessage}
\nocite{veličković2022message}
\nocite{rong2020dropedge}
\nocite{balcilar2021breaking}
\nocite{gilmer2017neural}
\nocite{GNNBook-ch12-guo}
\nocite{weisfeiler1968reduction}
\nocite{Huang_2021}
\nocite{nikolentzos2020k}
\nocite{wang2024empiricalstudyrealizedgnn}
\nocite{maron2020provably}
\nocite{NEURIPS2023_ebf95a6f}
\nocite{hwang2022an}
\nocite{zhang2024rethinking}
\nocite{GNNBook-ch5-li}
\nocite{tang2023chebnet}
\nocite{abboud2021surprising}
\nocite{thakoor2023largescalerepresentationlearninggraphs}
\nocite{papp2022theoreticalcomparisongraphneural}
\nocite{zhao2022starssubgraphsupliftinggnn}
\nocite{dwivedi2022benchmarkinggraphneuralnetworks}
\nocite{li2023distancematrixgeometricdeep}
\nocite{corso2020principalneighbourhoodaggregationgraph}
\nocite{zaheer2018deepsets}
\nocite{grossmann2023poster}

\newpage
\appendix
\section{Weisfeiler-Leman Test}
\label{appendix:WL}

\noindent\textbf{1-WL Test}
The 1-dimensional Weisfeiler-Leman (1-WL) test \cite{weisfeiler1968reduction} applies iterative color refinement to distinguish between non-isomorphic graphs. For each node \( v \in V \) in a graph \( G = (V, E) \), the state \( h_v^{(\ell)} \) of the node is refined iteratively, starting from initial labels (or colors) \( h_v^{(0)} \) based on the node's features. 

\noindent At iteration \( \ell \), the refinement is computed as follows:
\begin{equation} \textbf{\textit{1-WL}}:  h_v^{(\ell)} =\text{hash}\left(h_v^{(\ell-1)}, \{h_u^{(\ell-1)} : u \in \mathcal{N}(v)\}\right) \quad \forall v \in V
\end{equation}
where \( \mathcal{N}(v) \) represents the set of neighbors of node \( v \) in graph, and \( \text{hash} \) is a function that combines the current state of \( v \) with the states of its neighbors. 

\noindent The process repeats until the states of all nodes converge. If two graphs \( G \) and \( G' \) are non-isomorphic, then at least one node's state will differ, such that \( \text{WL}(G) \neq \text{WL}(G') \).

\begin{figure}[h!]
\centering
\resizebox{0.5\textwidth}{!}{  
\begin{tikzpicture}
    \node[circle, draw, fill=white] (a) at (0, 0) {1};
    \node[circle, draw, fill=white] (b) at (2, 0) {2};
    \node[circle, draw, fill=white] (c) at (0, 2) {3};
    \node[circle, draw, fill=white] (d) at (2, 2) {4};
    
    \draw (a) -- (b);
    \draw (a) -- (c);
    \draw (a) -- (d);
    \draw (b) -- (c);
    \draw (c) -- (d);
    
    \node[circle, draw, fill=white] (e) at (5, 0) {1};
    \node[circle, draw, fill=white] (f) at (7, 0) {2};
    \node[circle, draw, fill=white] (g) at (5, 2) {3};
    \node[circle, draw, fill=white] (h) at (7, 2) {4};
    
    \draw (e) -- (f);
    \draw (e) -- (g);
    \draw (f) -- (g);
    \draw (f) -- (h);
    \draw (g) -- (h);
\end{tikzpicture}
}  
\caption{Two simple graphs that cannot be distinguished by the 1-WL test.}
\label{1-WL}
\end{figure}
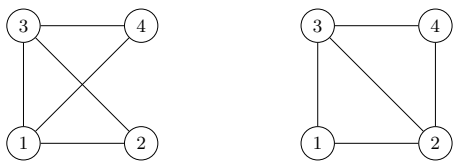

\noindent The WL algorithm effectively distinguishes the majority of graph pairs but it fails with certain fundamental examples. Figure \ref{1-WL} is a simple example. Although these graphs are non-isomorphic, they produce identical color sequences after the 1-WL refinement, making them indistinguishable by the test.

\noindent\textbf{Connection Between GNNs and 1-WL Test}
The following lemma is given by Morris et al.\,\cite{morris2021weisfeilerlemanneuralhigherorder} and Xu et al.\,\cite{xu2018powerful}.

\begin{mylemma}
    Let \( G_1 \) and \( G_2 \) be any two non-isomorphic graphs. If a graph neural network \( \mathcal{A} : G \to \mathbb{R}^d \) maps \( G_1 \) and \( G_2 \) to different embeddings, the Weisfeiler-Lehman graph isomorphism test also decides \( G_1 \) and \( G_2 \) are not isomorphic.
\end{mylemma}

\textbf{k-WL Test} \cite{li2023distancematrixgeometricdeep}
The \(k\)-dimensional Weisfeiler-Lehman test (\(k\)-WL) generalizes the classical test by assigning colors to \(k\)-tuples of nodes, denoted as \( v := (v_1, v_2, \dots, v_k) \in V^k \). Initially, \(k\)-tuples are labeled based on their isomorphism types, and these labels are iteratively refined according to the \(i\)-neighbors \( \mathcal{N}_i(v) \) of each tuple.
\begin{equation}
    \mathcal{N}_i(v) = \{(v_1, ..., v_{i-1}, u, v_{i+1}, ..., v_k) \mid u \in V\}
\end{equation}

To update the label of each tuple, k-WL iterates as defined:
\begin{equation} \textbf{\textit{k-WL}}:  h^{\ell}_{v} = \text{HASH}\left( h^{\ell-1}_{v}, \left\{ h^{\ell-1}_{u} \mid u \in \mathcal{N}_i(v) \right\} \mid i \in [k] \right).
\end{equation}

\section{EXP and BREC Datasets}
\label{appendix:datasets}

\textbf{BREC dataset}
consists 800 non-isomorphic graphs, organized into 400 pairs across four categories: Basics, Regular, Extension, and CFI.

\begin{itemize}
    \item [\textbullet] \textbf{Basics}: There are 60 pairs of 1-WL indistinguishable graphs, which can be regarded as a complement to EXP dataset with similar difficulty.
    \item [\textbullet] \textbf{Regular}: 140 pairs, including simple regular graphs, strongly regular graphs, 4-vertex condition graphs and distance regular graphs. All nodes of a regular graph possess the same degree and it is 1-WL indistinguishable.
    \item [\textbullet] \textbf{Extension}: Extension graphs consist of 100 pairs inspired by Papp and Wattenhofer \cite{papp2022theoreticalcomparisongraphneural} and generated and sampled by Wang et al. \cite{wang2024empiricalstudyrealizedgnn} between 1-WL and 3-WL distinguishing difficulty.
    \item [\textbullet] \textbf{CFI}: CFI graphs include 100 pairs. CFI graphs have larger graph sizes than other sections (up to 198 nodes) and can reach to 4-WL indistinguishable.
\end{itemize}

\begin{figure}[htbp]
    \centering
    \begin{subfigure}[b]{0.9\textwidth}
        \centering
        \includegraphics[width=\textwidth]{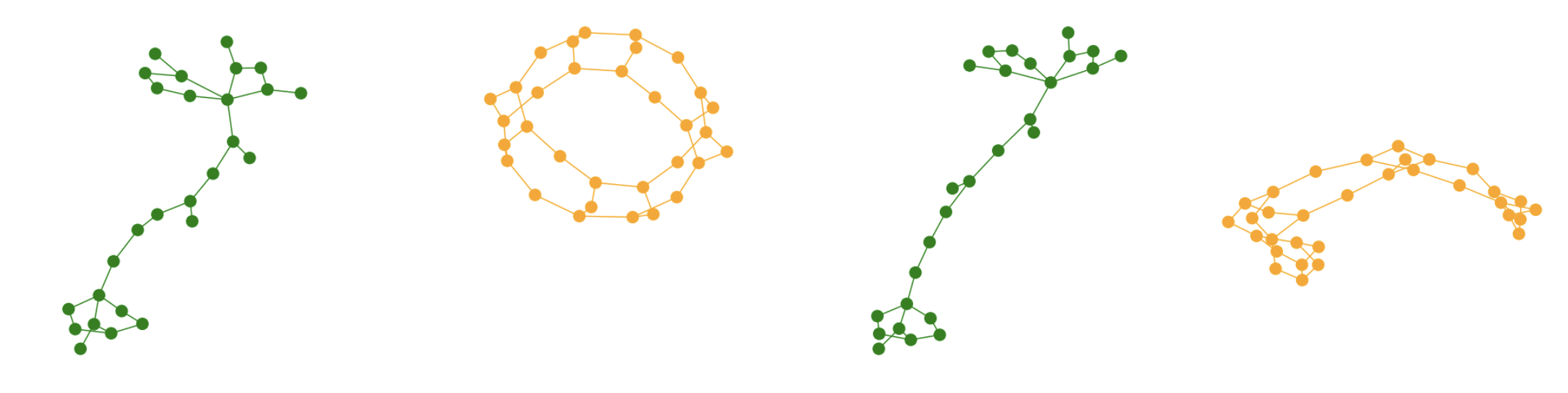}
        \caption{EXP dataset: \textcolor{orange}{orange} - core pair, \textcolor{mygreen}{green} - planar component}
        \label{fig:sub1z}
    \end{subfigure}
    
    \begin{subfigure}[b]{0.45\textwidth}
        \centering
        \includegraphics[width=\textwidth]{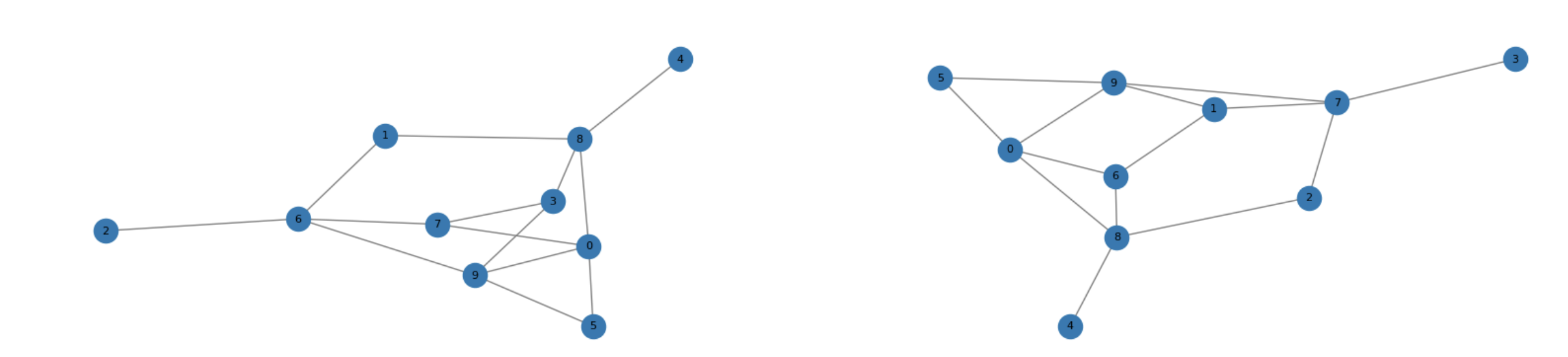}
        \caption{BREC-Basic}
        \label{fig:sub2x}
    \end{subfigure}
    \hfill
    \begin{subfigure}[b]{0.45\textwidth}
        \centering
        \includegraphics[width=\textwidth]{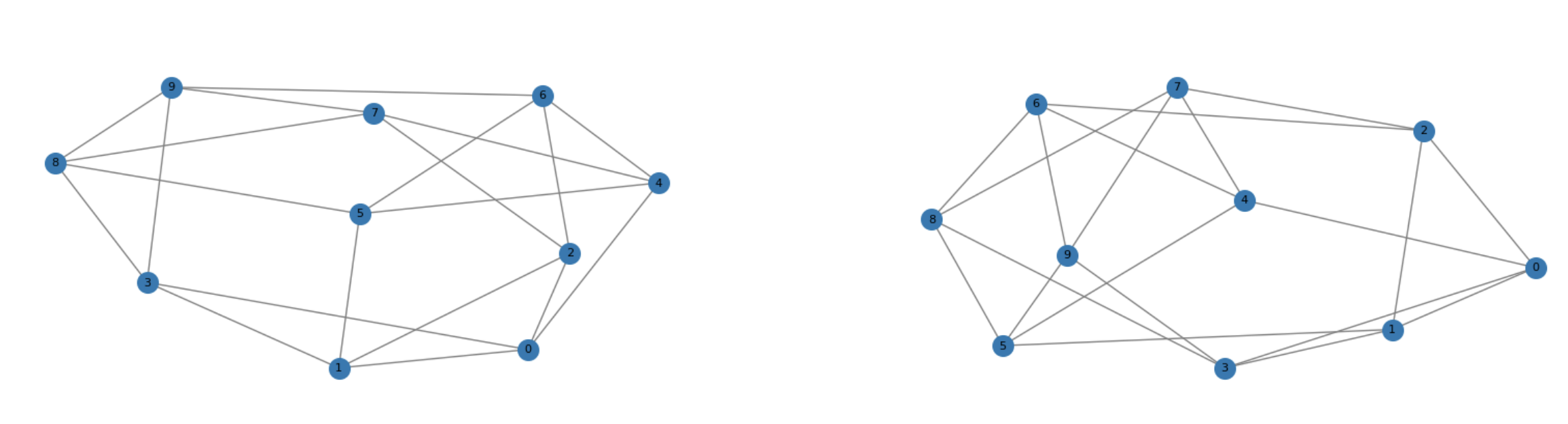}
        \caption{BREC-Regular}
        \label{fig:sub3x}
    \end{subfigure}
    
    \begin{subfigure}[b]{0.45\textwidth}
        \centering
        \includegraphics[width=\textwidth]{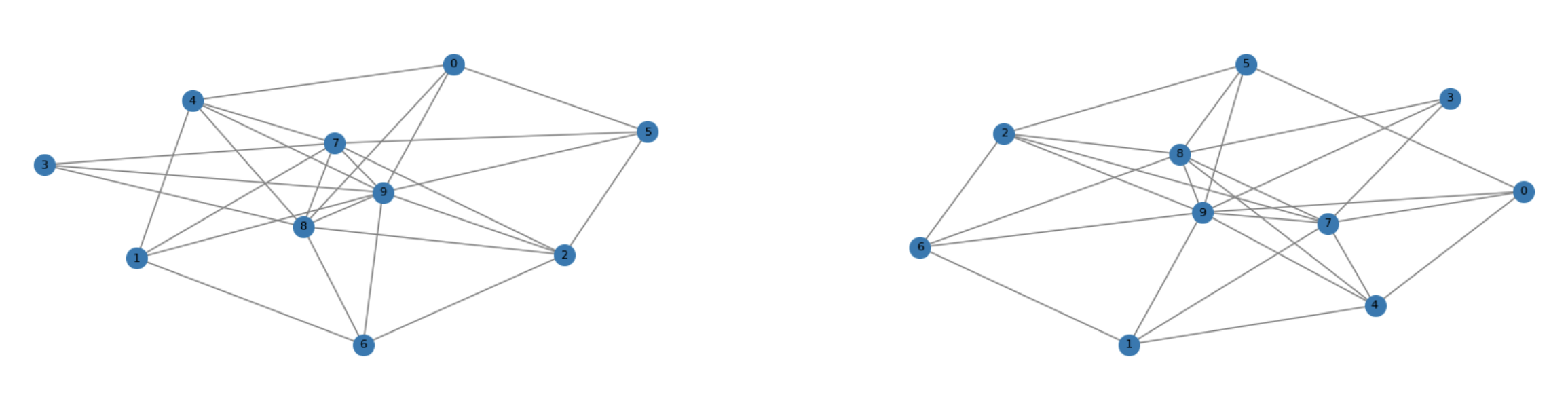}
        \caption{BREC-Extension}
        \label{fig:sub4x}
    \end{subfigure}
    \hfill
    \begin{subfigure}[b]{0.45\textwidth}
        \centering
        \includegraphics[width=\textwidth]{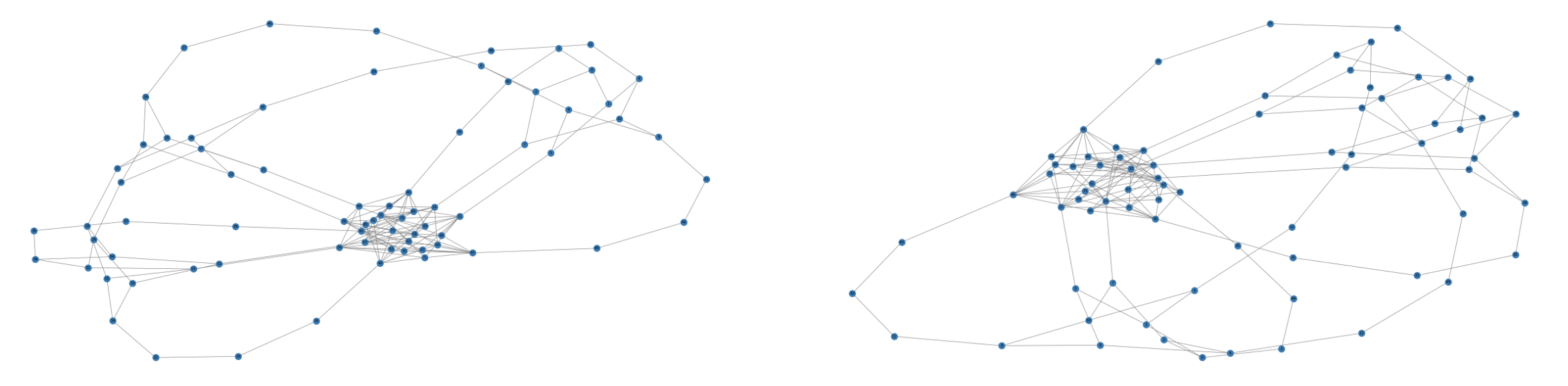}
        \caption{BREC-CFI}
        \label{fig:sub5x}
    \end{subfigure}
    
    \begin{subfigure}[b]{0.45\textwidth}
        \centering
        \includegraphics[width=\textwidth]{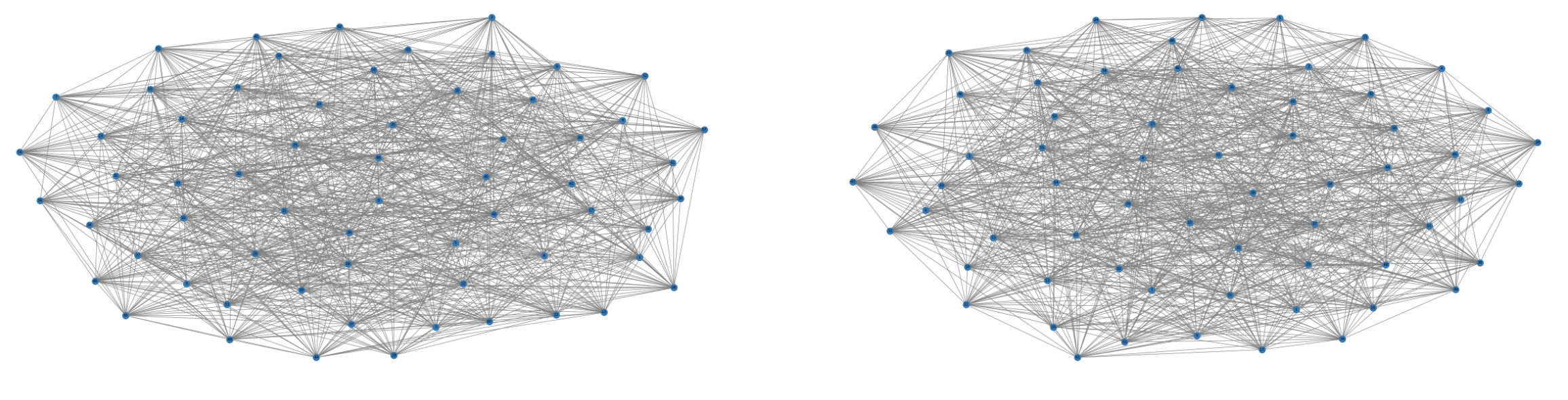}
        \caption{BREC-4-Vertex\_Condition}
        \label{fig:sub6x}
    \end{subfigure}
    \hfill
    \begin{subfigure}[b]{0.45\textwidth}
        \centering
        \includegraphics[width=\textwidth]{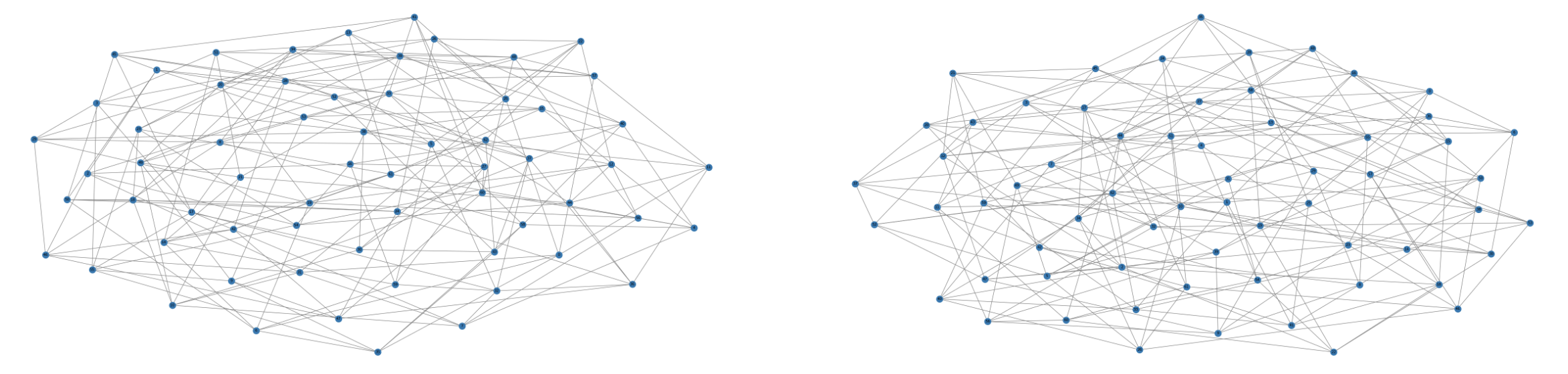}
        \caption{BREC-Distance Regular}
        \label{fig:sub7x}
    \end{subfigure}
    \caption{Dataset samples}
    \label{fig:main}
\end{figure}

Samples of EXP and BREC datasets are shown in Figure\ref{fig:main}

\section{Model Selection: Graph Isomorphism Networks, Principal Neighbourhood Aggregation, Deep Set}
\label{appendix:GNNs}

\noindent\textbf{Graph Isomorphism Network (GIN)}
GIN was proposed by Xu et al.\cite{xu2018powerful} which generalizes the WL test and hence achieves maximum discriminative power among GNNs and is as powerful as the 1-WL test. GIN updates node representations as:
\begin{align} \label{GIN-agg}
h_v^{(k)} =   {\rm MLP}^{(k)}   \left(  \left( 1 + \epsilon^{(k)} \right) \cdot h_v^{(k-1)} +  \sum\nolimits_{u \in \mathcal{N}(v)} h_u^{(k-1)}\right).
\end{align}
where $\epsilon^{(k)}$ is a newly introduced learnable parameter used to adjust the weight of the central node.

\noindent\textbf{Principal Neighbourhood Aggregation(PNA)}
PNA was proposed by Corso et al. \cite{corso2020principalneighbourhoodaggregationgraph} which combines multiple aggregators with degree scalers to capture the diversity if structural roles of nodes in a graph. 

The PNA layer updates the node feature \({h}_v\) of node \(v\) as follows:
\begin{equation}
    {h}_v^{(l+1)} = \gamma^{(l)} \left(h_v^{(l)}, \text{AGGREGATORS}\left( \left\{ h_u^{(l)}: u \in \mathcal{N}(v) \right\} \right) \right)
\end{equation}
where AGGREGATORS include \textit{mean}, \textit{sum}, \textit{max}, \textit{min} and \textit{std} aggregation methods.

\noindent\textbf{DeepSets(DS)} DeepSets, proposed by Zaheer et al. \cite{zaheer2018deepsets}, can operate on sets: a sequence of input features, ensuring that the output is invariant to the permutation of the input set elements. 

Given a set \( X = \{x_1, x_2, ..., x_n\} \), where each \( x_i \in \mathbb{R}^d \), we define a DeepSet neural network as:
\begin{equation}
    y = \text{MLP}_2 \left( \sum_{x_i \in X} \text{MLP}_1(x_i) \right),
\end{equation}
where \(\text{MLP}_1\) maps each element \( x_i \) of the set to a latent representation of size \( h \) and \(\text{MLP}_2\) takes the aggregated representation and maps it to the final output.

\section{Appendix D: Experiment Setup}
\label{appendix:setup}

\textbf{Dataset Format} EXP dataset is generated and stored using Pytorch Geometric. BREC dataset can be stored and read in NetworkX library.  To regulate and simplify the test, we store all datasets using Deep Graph Library (DGL). The conversion between libraries (PyTorch Geometric, NetworkX, and DGL) was handled carefully to ensure consistency in graph structures and node features.

\textbf{Graph Transformations} For the \textit{Virtual Node} and \textit{Graph Encoding} transformations, we utilized pre-built functions from the \textbf{torch\_geometric.transforms} module. Specifically, we applied \texttt{VirtualNode} to add a global node connected to all other nodes, and \texttt{AddLaplacianEigenvectorPE} to compute and append the top eigenvectors of the Laplacian matrix as positional encodings to the node features. For methods such as \textit{Centrality}, \textit{Distance Encoding}, and \textit{Subgraph Extraction}, we used the NetworkX library to compute additional node features. These features, calculated based on graph-theoretic measures, were concatenated with the original node features before being passed to the GNN models. Subgraph extraction was performed by generating ego-graphs (local subgraphs) within a fixed radius of each node.

\textbf{GNN Architectures} We experimented GIN, PNA and DS. The GNN architectures were implemented using 4 layers, with the following hyperparameters: \begin{itemize} 
\item [\textbullet] \textbf{Learning rate:} 
0.001
\item [\textbullet] \textbf{Optimizer:} Adam optimizer

\item [\textbullet] \textbf{Batch size:} 32 
\item [\textbullet] \textbf{Epochs:} 100 

\end{itemize}

Each GNN model was trained using a contrastive learning framework to acquire different output embeddings for non-isomorphic pairs and same graph embeddings for isomorphic pairs.

\end{document}